\documentclass[10pt,twocolumn,letterpaper]{article}

\usepackage{cvpr}              

\usepackage{amsmath}
\usepackage{amssymb}
\usepackage{array}
\usepackage{bm}
\usepackage{caption}
\usepackage{siunitx}

\definecolor{cvprblue}{rgb}{0.21,0.49,0.74}
\usepackage[pagebackref,breaklinks,colorlinks,allcolors=cvprblue]{hyperref}
\usepackage{multirow}

\def\paperID{22} 
\def\confName{CVPR}
\def\confYear{2026}

\title{SSFT: A Lightweight Spectral–Spatial Fusion Transformer \\ for Generic Hyperspectral Classification}

\author{
Alexander Musiat$^{1}$ \qquad
Nikolas Ebert$^{1}$ \qquad
Oliver Wasenm\"uller$^{1}$ 
\\
$^{1}$Mannheim University of Applied Sciences, Germany\\
{\tt\small \{a.musiat, n.ebert, o.wasenmueller\}@th-mannheim.de}}

\begin{document}
\twocolumn[{
\maketitle
\begin{center}
    \captionsetup{type=figure}
    \includegraphics[width=\textwidth]{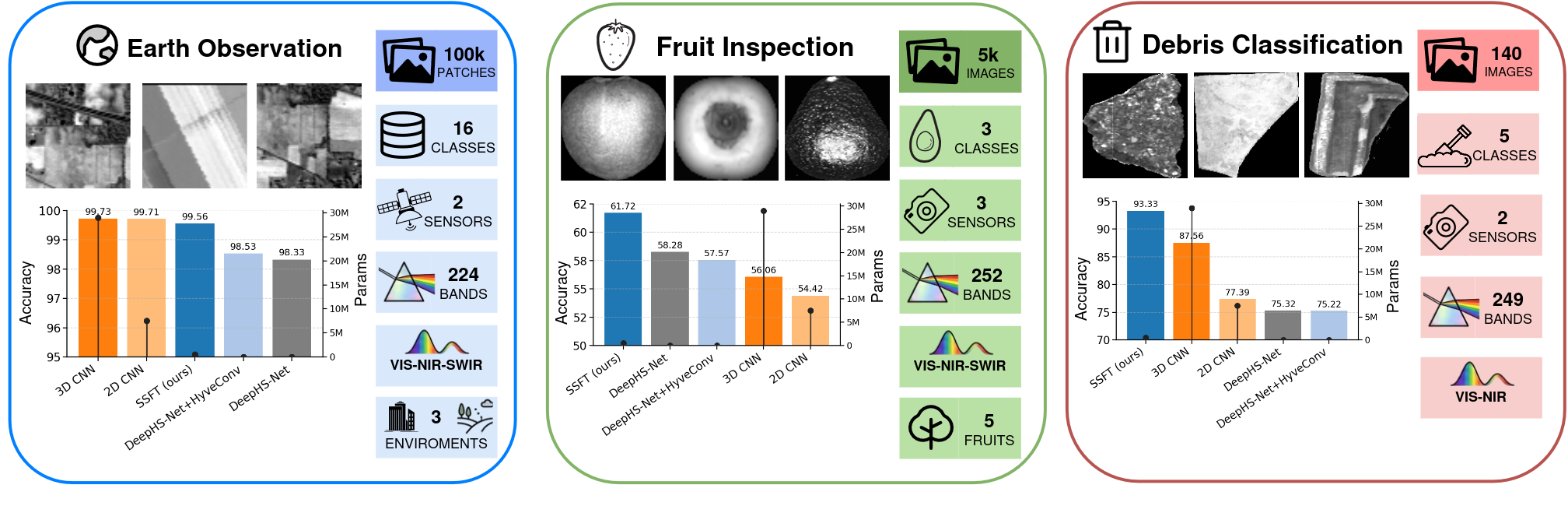}
    \captionof{figure}{
    To demonstrate the general applicability of SSFT, we evaluate on the heterogeneous sub-datasets of the HSI-Benchmark \cite{frank2023_hsibenchmark}, covering earth observation, fruit condition assessment, and fine-grained debris material recognition. The figure shows representative image patches and key dataset characteristics for each domain. We further compare the top-performing methods \cite{paoletti2019_mlp_1dcnn_2dcnn_3dcnn, varga2021_deephsnet, varga2023_deephsnetplushyveconv} per domain in terms of accuracy (bars, left axis) and model size (lollipops, right axis). Across this diverse multi-domain protocol, SSFT achieves state-of-the-art overall performance, ranking first on fruit and debris while remaining competitive on earth observation despite its compact architecture using less than 2\% of the parameters of the previous leading method.
    }\label{fig:teaser}
\end{center}
}]
\begin{abstract}
Hyperspectral imaging enables fine-grained recognition of materials by capturing rich spectral signatures, but learning robust classifiers is challenging due to high dimensionality, spectral redundancy, limited labeled data, and strong domain shifts. Beyond earth observation, labeled HSI data is often scarce and imbalanced, motivating compact models for generic hyperspectral classification across diverse acquisition regimes. We propose the lightweight Spectral-Spatial Fusion Transformer (SSFT), which factorizes representation learning into spectral and spatial pathways and integrates them via cross-attention to capture complementary wavelength-dependent and structural information. We evaluate our SSFT on the challenging HSI-Benchmark, a heterogeneous multi-dataset benchmark covering earth observation, fruit condition assessment, and fine-grained material recognition. SSFT achieves state-of-the-art overall performance, ranking first while using less than 2\% of the parameters of the previous leading method. We further evaluate transfer to the substantially larger SpectralEarth benchmark under the official protocol, where SSFT remains competitive despite its compact size. Ablation studies show that both spectral and spatial pathways are crucial, with spatial modeling contributing most, and that SSFT remains robust without data augmentation.
\end{abstract}    
\section{Introduction}
\label{sec:introduction}

Hyperspectral imaging (HSI) enables fine-grained discrimination of materials and physical states by capturing rich spectral signatures for each pixel across dozens to hundreds of narrow bands. This makes HSI a powerful sensing modality for applications such as earth observation, agriculture, environmental monitoring, and quality assessment. At the same time, HSI datasets are inherently diverse: measurements vary across sensors and optics, spectral coverage and band spacing, spatial resolution, illumination conditions, and calibration and preprocessing, creating substantial domain shifts across application regimes. Moreover, beyond earth observation, annotated HSI datasets are typically small and class-imbalanced, limiting supervision and amplifying overfitting in real-world deployments. 

Most modern HSI classifiers build on deep backbones adapted from RGB image recognition \cite{Ahmad2021_hsi_survey_3}. Early work \cite{li2017_spectral_3dcnn_early_1, he2017_multiscale_3dcnn_early_2} largely relied on 3D convolutions to jointly model spectral-spatial correlations, while later approaches employed 2D CNNs with spectral mixing \cite{zheng2020_hyperspectral_2dcnn_spectral_mixing}, attention mechanisms \cite{hang2020_hyperspectral_2dcnn_attention}, or transformer encoders \cite{he2019hsi_bert_transformer, hong2021_spectralformer} to capture wider context. This progression has improved representational power, but it has also shifted the field toward increasingly large and computationally expensive architectures. While such models can perform well when labeled data is abundant and train/test conditions are closely matched, these assumptions often do not hold in HSI. In diverse low-data regimes, high-capacity models may instead exploit dataset-specific statistics or sensor and preprocessing artifacts rather than transferable spectral-spatial cues. Moreover, augmentation strategies commonly used in RGB vision do not transfer directly to HSI, since arbitrary spectral perturbations can violate acquisition physics and introduce unrealistic shortcuts.

To address these challenges, we propose the Spectral-Spatial Fusion Transformer (SSFT), a compact architecture tailored to heterogeneous, low-supervision HSI regimes. SSFT processes spectral signatures and spatial structure in separate lightweight branches and integrates them through cross-attention, enabling controlled interaction between complementary cues without resorting to large backbones. With only 516k parameters, SSFT is orders of magnitude smaller than common CNN and transformer baselines \cite{paoletti2019_mlp_1dcnn_2dcnn_3dcnn, yang2022_hit}, yet achieves the best overall score on HSI-Benchmark \cite{frank2023_hsibenchmark} (see Fig.~\ref{fig:teaser}), ranking first on Fruit and Debris while remaining competitive on earth observation. We further evaluate transfer to the substantially larger SpectralEarth benchmark \cite{Braham2025_spectralearth_dataset} under the official protocol, where SSFT remains competitive despite its compact size. Finally, we complement the main results with controlled branch ablations, auxiliary supervision experiments, and an analysis of data augmentation, showing that both branches are important and that robustness in HSI cannot be assumed from RGB training recipes.
\section{Related Works}
\label{sec:related_works}


\subsection{Hyperspectral Image Classification}
\label{subsec:hyperspectral_image_classification}

Hyperspectral images are naturally represented as a spectral-spatial data cube, where each pixel is not a three-channel color vector but a densely sampled reflectance spectrum measured across dozens to hundreds of narrow wavelengths. These spectra often carry subtle absorption patterns tied to material composition and physical condition, but they are also highly structured: neighboring bands are strongly correlated, signal-to-noise varies across the spectrum, and measurements depend on acquisition factors such as optics, illumination, and calibration. As a result, hyperspectral classification is not merely a higher-dimensional version of RGB recognition; it is a data-efficient learning problem that must extract transferable spectral cues while exploiting spatial context, typically under limited and costly supervision \cite{kumar2024_hsi_survey_1, ganji2024_hsi_survey_2, Ahmad2021_hsi_survey_3}.

A large fraction of widely used hyperspectral benchmarks stems from satellite earth observation, where a small set of canonical scenes has historically driven progress and where more recent releases move toward larger-scale, multi-scene evaluation \cite{indianpines_salinas_pavia_ehu,debes2014_houston2013_dataset, Braham2025_spectralearth_dataset}. However, beyond earth observation, many close-range hyperspectral applications in areas such as agriculture, food quality inspection, or environmental sensing rely on datasets that are typically smaller, more class-imbalanced, and collected under varied acquisition setups, which amplifies domain shift and makes results harder to compare across studies \cite{Gaidel2023_AgriculturalPlantHSI_small_1, Ortega2024_HistologyHSIGB_small_2}. To better reflect these real-world conditions, recent efforts have introduced multi-domain benchmarks that combine datasets from distinct application regimes and enforce standardized protocols, enabling systematic assessment of transfer across domains rather than performance tuned to a single sensor or scene \cite{frank2023_hsibenchmark,cherif2025_greenhyperspectra}.

Deep hyperspectral classifiers can be grouped by where they mix information in the cube. Spectral models treat each pixel spectrum as a 1D signal and learn cross-band patterns along the wavelength axis \cite{zeng2021_spectral1dcnn,pourdarbani2021_nitrogen1dcnn}. They are data-efficient and preserve fine spectral detail, but largely ignore spatial context that is often essential for disambiguating classes. Spatial CNN pipelines instead apply 2D convolutions on spatial patches, handling the spectral dimension either via projected cube representations (\eg, band selection or PCA channels) or via learned spectral mixing (\eg, $1{\times}1$ convolutions) before 2D processing, thereby leveraging texture and local structure \cite{yang2018_hyperspectral2dcnn,paoletti2020_rotation2dcnn}. 3D CNNs avoid this trade-off by convolving directly over spectral–spatial patches \cite{zhou2022_spectralfusion3dcnn,jung2022_graymold3dcnn,firat2023_residualspatialspectral3dcnn}, but their dense joint processing is computationally heavy and can overfit small benchmarks. Attention-based models provide an alternative route to capture non-local spectral-spatial dependencies \cite{hong2021_spectralformer, chakraborty2021_spectralnet}. However, applying self-attention to high-dimensional hyperspectral tokens can be similarly costly, and vanilla transformer backbones are often too parameter- and compute-intensive for small, heterogeneous HSI regimes. Several recent works therefore combine spectral and spatial modeling more explicitly through dedicated fusion mechanisms. For example, spectral-spatial transformer architectures have been proposed to jointly encode both modalities and enhance cross-dimensional interaction \cite{Liao2023_spectral_spatial_fusion_transformer, Sun2024_massformer}, while lightweight fusion networks combine separate spectral and spatial feature extractors followed by feature fusion to improve efficiency \cite{Chen2020_lightweight_spectral_spatial_fusion}.

Across these model families, the recurring tension is between expressive spectral–spatial integration and robustness under limited supervision. This tension is amplified in diverse multi-domain settings, where larger backbones can latch onto dataset-specific statistics and where training recipes do not transfer uniformly across domains. These observations motivate compact, factorized designs with controlled interaction mechanisms and careful empirical evaluation under heterogeneous hyperspectral conditions.


\begin{figure*}[t]
  \centering
  \includegraphics[width=\textwidth]{images/method_overview-1.png}
  \caption{Overview of the proposed SSFT architecture. Given an input hyperspectral data cube, the network factorizes representation learning into two complementary pathways: a spatial encoder that captures local image structure via lightweight 2D convolutions and pooling, and a spectral encoder that models wavelength-dependent signatures by embedding spectral information and applying self-attention. The resulting spatial and spectral features are aligned in a shared embedding space and fused in a cross-attention fusion module, enabling bidirectional interaction between spatial context and spectral cues. The fused representation is pooled and passed to the main classifier head to produce the final prediction. During training, two auxiliary classification heads provide additional supervision for the individual branches; these heads are discarded during inference and do not affect test-time complexity.}
  \label{fig:wide}
\end{figure*}

\subsection{Augmentation and Deep Supervision in HSI}
\label{subsec:augmentation_and_deep_supervision_in_hsi}

Performance in hyperspectral image classification is often limited not only by model capacity, but also by supervision constraints: labeled datasets are typically small, class distributions can be imbalanced, and acquisition conditions vary across sensors and setups \cite{jia2021_hsi_fewlabel_survey, signoroni2021_dl_meets_hsi, zheng2022_imbalanced_hsi}. Training strategies therefore matter in practice. In RGB vision, data augmentation is a standard tool to improve invariance and robustness as well as reduce overfitting \cite{ebert2025enhancing, oehri2024genformer}. In HSI, however, augmentations must respect measurement physics: hyperspectral signatures are shaped by illumination, sensor response, calibration, and noise characteristics, and arbitrary spectral perturbations can create unrealistic signals that undermine transferable cues \cite{khonina2024_ai_hsi_synergy}.

HSI augmentations can be broadly grouped into spatial and spectral transformations. Spatial augmentations (e.g., flipping, rotation, cropping) regularize spatial feature learning and improve robustness to geometric variability, whereas spectral augmentations operate along the band dimension and include additive noise, band-wise gain/offset perturbations, spectral masking or band dropping, and band-block perturbations; related approaches mimic illumination changes via intensity scaling or shading patterns. While such transformations can act as effective regularizers, their benefit depends on whether they reflect plausible variation sources for the target sensors and domains \cite{nalepa2019_hyperspectral_data_augmentation_1, li2018_data_augmentation_hsi_cnn_2}. This is particularly critical in multi-domain benchmarks, where an augmentation that is helpful in one regime may distort class-discriminative spectra in another, motivating careful empirical validation rather than direct reuse of RGB heuristics. A complementary line of work improves optimization and representation learning through additional supervision signals \cite{luo2022_deeply_supervised_pseudo_learning_hsi, li2019_hmcnn_ac_hsi_supervision}. Deep supervision attaches auxiliary classifier heads to intermediate features and trains them jointly with the final output, encouraging informative intermediate representations and providing shorter gradient paths. This can stabilize training and reduce overfitting, especially when data are scarce and optimization is sensitive to hyperparameters. These considerations motivate systematic evaluation of augmentation choices and auxiliary supervision under broad, multi-domain HSI protocols.
\section{Method}
\label{sec:method}

A hyperspectral image is represented as a data cube $X~\in~\mathbb{R}^{H \times W \times C}$, where $H$ and $W$ denote the spatial resolution and $C$ the number of spectral bands. Given $X$, our goal is to predict a class label $y \in \{1,\ldots,K\}$. SSFT factorizes spectral and spatial processing into two parallel branches. The spectral encoder $\mathcal{E}_s$ extracts a representation $h^{(s)} = \mathcal{E}_s(X)$ that focuses on correlations across bands, while the spatial encoder $\mathcal{E}_p$ produces $h^{(p)} = \mathcal{E}_p(X)$ capturing local spatial context. The representations are combined with a fusion module $h = \Phi\!\left(h^{(s)}, h^{(p)}\right)$ and the fused features are then fed into a classifier $g$ to obtain logits $z = g(h)$ and the prediction $\hat{y} = \arg\max_k z_k$. The spectral and spatial encoders are described in Sec.~\ref{subsec:spectral_encoder} and Sec.~\ref{subsec:spatial_encoder}, respectively, followed by the fusion module in Sec.~\ref{subsec:fusion_module}, the classification heads in Sec.~\ref{subsec:classification_heads} and the training objective in Sec.~\ref{subsec:training_objective}. Lastly, the implementation details are described in Sec.~\ref{subsec:implementation_details}.


\subsection{Spectral Encoder}
\label{subsec:spectral_encoder}

The spectral encoder $\mathcal{E}_s$ models inter-band dependencies by applying self-attention along the spectral dimension independently at each spatial location. This design is motivated by the strong correlation structure of neighboring spectral bands and the need to model longer-range inter-band dependencies in a content-adaptive manner, for which attention-based spectral modeling has proven effective in prior HSI work \cite{hong2021_spectralformer,Liao2023_spectral_spatial_fusion_transformer}.

Given an input data cube $X \in \mathbb{R}^{H \times W \times C}$, we first downsample the spatial resolution by a factor $s$ using max-pooling, yielding $X^{(s)\downarrow} \in \mathbb{R}^{\frac{H}{s} \times \frac{W}{s} \times C}$. We then flatten the spatial grid into $N = \frac{H}{s}\cdot\frac{W}{s}$ locations and form, for each location, a spectral token sequence of length $C$, denoted by $\tilde{X}^{(s)} \in \mathbb{R}^{N \times C}$.
A learnable embedding function $\phi(\cdot)$ maps each scalar band value to a $D$-dimensional token embedding, yielding $E = \phi(\tilde{X}^{(s)}) \in \mathbb{R}^{N \times C \times D}$. To encode band identity, we add a learned channel embedding $P \in \mathbb{R}^{C \times D}$ to the token embeddings, yielding $Z = E + P$, where $P$ is broadcast across the $N$ spatial locations. We then apply a single spectral self-attention block to model inter-band interactions. Queries, keys, and values are obtained via learned linear projections, $Q = ZW_Q$, $K = ZW_K$, and $V = ZW_V$, with $W_Q, W_K, W_V \in \mathbb{R}^{D \times D}$. Scaled dot-product attention is computed over the $C$ spectral tokens:
\begin{equation}
\mathrm{Attn}(Z) = \mathrm{softmax}\!\left(\frac{QK^\top}{\sqrt{D}}\right)V,
\label{eq:spectral_attn}
\end{equation}
where the softmax is applied along the channel dimension. The attention output is followed by residual connections and a position-wise two-layer Feed-Forward Network (FFN). In practice, we implement the spectral encoder with multi-head self-attention, following established conventions \cite{dosovitskiy2021vit, ebert2023plg}.

Finally, we aggregate the $C$ channel tokens into a single feature vector per spatial location via mean pooling and reshape them back into a feature map:
\begin{equation}
h^{(s)} \in \mathbb{R}^{\frac{H}{s} \times \frac{W}{s} \times D}.
\label{eq:spectral_out}
\end{equation}
The resulting features are forwarded to the fusion module.


\subsection{Spatial Encoder}
\label{subsec:spatial_encoder}

The spatial encoder $\mathcal{E}_p$ extracts features that emphasize local spatial structure \eg, edges, textures, and region-level patterns. While the spectral encoder mixes information across bands at each spatial location, the spatial encoder aggregates information across neighboring pixels in the image plane using convolutional processing. We adopt a lightweight convolutional design because local spatial patterns such as edges, textures, and region structure are well captured by CNN inductive biases, while keeping the spatial branch compact and data-efficient \cite{paoletti2019_mlp_1dcnn_2dcnn_3dcnn,he2016_resnet}.

Given an input data cube $X \in \mathbb{R}^{H \times W \times C}$, we first project the $C$ spectral channels to the shared embedding dimension $D$ using a $1 \times 1$ convolution, yielding $X^{(p)} = \psi(X) \in \mathbb{R}^{H \times W \times D}$. To reduce computational cost and to match the spatial resolution of the spectral branch, we then downsample the feature map by a factor $s$ via max-pooling, resulting in $X^{(p)\downarrow} \in \mathbb{R}^{\frac{H}{s} \times \frac{W}{s} \times D}$.
We then apply a single convolutional block to obtain the spatial representation:
\begin{equation}
h^{(p)} \in \mathbb{R}^{\frac{H}{s} \times \frac{W}{s} \times D}.
\label{eq:spatial_out}
\end{equation}
The convolutional block consists of a $3 \times 3$ convolution, followed by batch normalization and a GELU non-linearity. This design provides inductive biases toward locality and translation equivariance, making it effective for modeling spatial patterns in hyperspectral imagery, including edges, textures, and region-level structures. The resulting feature map $h^{(p)}$ is forwarded to the fusion module.

\subsection{Fusion Module}
\label{subsec:fusion_module}

The fusion module $\Phi$ integrates spectral cues into spatial representations in a selective, content-dependent manner. Given spectral features $h^{(s)}$ and spatial features $h^{(p)}$ at matched resolution, we flatten both feature maps into token sequences of length $N=\frac{H}{s}\cdot\frac{W}{s}$, yielding $S \in \mathbb{R}^{N \times D}$ and $P \in \mathbb{R}^{N \times D}$.
Here, $S$ represents per-location spectral descriptors produced by the spectral branch, while $P$ encodes spatial context features at the same locations. We employ cross-attention to inject spectral information into the spatial stream, letting spatial tokens act as queries and spectral tokens provide keys and values. This choice preserves the spatial inductive bias of the convolutional branch while enabling adaptive, location-specific spectral conditioning.

Concretely, queries, keys, and values are obtained via learned linear projections, $Q_p = PW_Q$, $K_s = SW_K$, and $V_s = SW_V$, with $W_Q, W_K, W_V \in \mathbb{R}^{D \times D}$. Cross-attention computes a weighted combination of spectral values for each spatial token:
\begin{equation}
\mathrm{CA}(P,S) = \mathrm{softmax}\!\left(\frac{Q_pK_s^\top}{\sqrt{D}}\right)V_s,
\label{eq:cross_attn}
\end{equation}
where the softmax is applied over the key dimension. Similar to the self-attention block in the spectral encoder, the cross-attention output is followed by residual connections and a position-wise two-layer FFN. In practice, we use multi-head cross-attention. The output sequence is reshaped back to a feature map and passed to the main classification head.


\subsection{Classification Heads}
\label{subsec:classification_heads}

Given the fused feature map, the main classification head first aggregates information across spatial locations via pooling and then predicts class logits with a lightweight MLP. In our default configuration, we use adaptive average pooling to obtain a single $D$-dimensional vector per sample. The pooled feature is passed through a two-layer MLP with GELU activation, producing logits $z \in \mathbb{R}^{K}$ for the final cross-entropy loss.

To stabilize optimization and encourage task-relevant intermediate representations, we optionally attach auxiliary classifier heads to the outputs of the spectral and spatial branches before fusion. Each auxiliary head follows the same design as the main head (pooling + MLP) and produces logits $z^{(s)}, z^{(p)} \in \mathbb{R}^{K}$. Auxiliary heads are used only during training; during inference, they are discarded and only the main head is used to predict the final class logits.


\subsection{Training Objective}
\label{subsec:training_objective}

We train SSFT using standard multi-class cross-entropy loss on the main classifier logits $z \in \mathbb{R}^{K}$ produced from the fused representation and denote this loss as $\mathcal{L}_{\mathrm{main}}$. To stabilize optimization and encourage task-relevant branch representations, we optionally apply deep supervision via auxiliary classifier heads attached to the spectral and spatial branch outputs prior to fusion. Let $z^{(s)}$ and $z^{(p)}$ denote the corresponding auxiliary logits, and let $\mathcal{L}^{(s)}_{\mathrm{aux}}$ and $\mathcal{L}^{(p)}_{\mathrm{aux}}$ be the respective cross-entropy losses. We optimize the weighted sum $\mathcal{L} = \mathcal{L}_{\mathrm{main}} + \lambda_{\mathrm{aux}}\left(\mathcal{L}^{(s)}_{\mathrm{aux}} + \mathcal{L}^{(p)}_{\mathrm{aux}}\right)$, where $\lambda_{\mathrm{aux}} \ge 0$ controls the strength of deep supervision.

\subsection{Implementation Details}
\label{subsec:implementation_details}

We implement SSFT in PyTorch. Unless stated otherwise, we use an embedding dimension of $D=64$ and downsample the spatial resolution by a factor of $s=8$ in both branches. The spectral encoder consists of one spectral self-attention block with four heads. The spatial encoder uses a single convolutional block consisting of a $3 \times 3$ convolution followed by batch normalization and GELU. The fusion module uses one cross-attention block with four heads. We train for up to 50 epochs with AdamW (initial learning rate $4\times10^{-4}$, weight decay $10^{-2}$, batch size 8). We employ a StepLR schedule that decays the learning rate by a factor of 0.1 every 20 epochs and apply early stopping with a patience of 10 epochs based on validation accuracy. The primary objective is multi-class cross-entropy on the main classifier outputs; when auxiliary heads are enabled, we optimize the weighted sum described in Sec.~\ref{subsec:training_objective}. Unlike many hyperspectral pipelines, we do not apply dimensionality reduction such as PCA, enabling fully end-to-end training.
\section{Experiments}
\label{sec:experiments}

To demonstrate the robustness of SSFT across domains, we evaluate it on diverse datasets spanning multiple application areas, including earth observation, fruit condition assessment, and fine-grained material recognition, with varying characteristics and acquisition devices.

The HSI-Benchmark \cite{frank2023_hsibenchmark} is a collection of multiple hyperspectral datasets spanning three substantially different hyperspectral application domains (HRSS, Fruit, Debris) with distinct acquisition settings and recognition challenges. HRSS consists of satellite earth observation scenes with  large spatial extent, exhibiting strong spatial structure and comparatively coarse material categories. Fruit is captured in close-range settings and focuses on condition/ripeness recognition, where classes often differ by subtle spectral cues while spatial diversity is limited. Debris contains visually similar materials and poses a fine-grained recognition problem that relies strongly on spectral discrimination to separate closely related material classes.

SpectralEarth \cite{Braham2025_spectralearth_dataset} is a large-scale hyperspectral earth observation benchmark that supports multiple downstream evaluation tasks under a unified protocol. In this work, we follow the official setup for the EnMAP-CORINE land-cover task, which formulates land-cover prediction as multi-label classification over large earth observation scenes. Since SSFT is designed for the limited-data regime of HSI-Benchmark, we use SpectralEarth as an additional large-scale benchmark to assess transfer and scalability of this compact architecture. This setting complements the heterogeneous, small-scale regimes in HSI-Benchmark by providing a substantially larger earth observation testbed with strong spatial structure and broad spectral variability, as well as a different label space and data distribution.

For HSI-Benchmark, we strictly follow the official evaluation protocol and report results for both settings defined by the benchmark (object-wise and patch-wise classification). We use the official train/validation/test splits, repeat all runs with three random seeds, and report test accuracy per domain as well as the benchmark’s overall mean score. Unless stated otherwise, inputs are normalized according to the benchmark protocol. For comparison, we report baseline results from the official benchmark under the same standardized splits, metrics, and evaluation settings.

For SpectralEarth, we follow the official two-stage training protocol and evaluate only the EnMAP-CORINE downstream task. We consider the benchmark’s standard backbone initializations (Random, MoCo-v2 and DINO, two common self-supervised pretraining schemes), use the official splits, and normalize inputs with the provided channel-wise statistics. Performance is reported as macro F1-score on the CORINE test split and averaged over three seeds. Comparisons are made against the benchmark’s standard spectral backbones under the same official downstream protocol.


\subsection{Results on HSI-Benchmark}
\label{subsec:main_results}

\begin{table}
\caption{Comparison of SSFT to prior methods on the HSI-Benchmark \cite{frank2023_hsibenchmark} test split. We report accuracy on HRSS, Fruit, and Debris, as well as the overall (mean) benchmark score, and include the number of trainable parameters for each model. Best results are shown in bold, and second-best results are underlined.}
\label{tab:benchmark_results}
\centering
\resizebox{\columnwidth}{!}{%
\begin{tabular}{@{}l
                S[table-format=2.2]
                cccc c@{}}
\toprule
Method & {Params (M)} & HRSS & Fruit & Debris & Overall \\
\midrule
RNN \cite{rumelhart1986_rnn} & 0.03 & 69.61 & 41.72 & 52.50 & 54.61 \\
MLP \cite{paoletti2019_mlp_1dcnn_2dcnn_3dcnn} & 0.03 & 71.18 & 44.54 & 50.72 & 55.48 \\
PLS-DA \cite{barker2003_plsda} & {\text{N/A}} & 66.61 & 51.22 & 38.56 & 52.13 \\
Attention-based CNN \cite{lorenzo2020_attentionbasedcnn} & 2.00 & 89.93 & 44.88 & 50.18 & 61.66 \\
ResNet-152 \cite{he2016_resnet} & 59.00 & 96.27 & 47.00 & 29.30 & 57.52 \\
ResNet-152+HyveConv \cite{frank2023_hsibenchmark} & 58.00 & 97.22 & 42.66 & 46.91 & 62.26 \\
SpectralFormer \cite{hong2021_spectralformer} & 1.00 & 96.25 & 41.71 & 53.24 & 63.74 \\
SVM \cite{cristianini2004_svm} & {\text{N/A}} & 78.85 & 47.10 & 52.57 & 59.50 \\
2D CNN (spectral) \cite{frank2023_hsibenchmark} & 7.50 & 86.73 & 49.27 & 50.47 & 62.15 \\
SpectralNET \cite{chakraborty2021_spectralnet} & 8.30 & 98.38 & 49.25 & 46.33 & 64.65 \\
2D CNN (spatial) \cite{paoletti2019_mlp_1dcnn_2dcnn_3dcnn} & 7.50 & 99.69 & 44.85 & 54.23 & 66.26 \\
HiT \cite{yang2022_hit} & 59.00 & 98.47 & 48.16 & 59.23 & 68.62 \\
1D CNN \cite{paoletti2019_mlp_1dcnn_2dcnn_3dcnn} & 0.07 & 91.04 & 51.30 & 63.13 & 68.49 \\
HybridSN \cite{roy2019_hybridsn} & 50.00 & 97.54 & 48.74 & 73.85 & 73.38 \\
ResNet-18 \cite{he2016_resnet} & 12.00 & 99.52 & 49.05 & 59.56 & 69.38 \\
EMP CNN \cite{ghamisi2018_gaborcnn_empcnn} & 7.50 & 99.54 & 52.76 & 61.87 & 71.39 \\
DeepHS-Hybrid-Net \cite{varga2023_deephshybridnet} & 1.30 & 95.36 & 55.01 & 82.14 & 77.50 \\
ResNet-18+HyveConv \cite{frank2023_hsibenchmark} & 11.00 & 99.67 & 51.43 & 67.12 & 72.74 \\
Gabor CNN \cite{ghamisi2018_gaborcnn_empcnn} & 7.40 & \textbf{99.75} & 52.57 & 66.43 & 72.92 \\
DeepHS-Net \cite{varga2021_deephsnet} & 0.03 & 98.33 & \underline{58.28} & 75.32 & 77.31 \\
DeepHS-Net+HyveConv \cite{varga2023_deephsnetplushyveconv} & 0.02 & 98.53 & 57.57 & 75.22 & 77.11 \\
2D CNN \cite{paoletti2019_mlp_1dcnn_2dcnn_3dcnn} & 7.50 & 99.71 & 54.42 & 77.39 & 77.17 \\
3D CNN \cite{paoletti2019_mlp_1dcnn_2dcnn_3dcnn} & 29.00 & \underline{99.73} & 56.06 & \underline{87.56} & \underline{81.12} \\
SSFT (ours) & 0.52 & 99.56 & \textbf{61.72} & \textbf{93.33} & \textbf{84.87} \\
\bottomrule
\end{tabular}%
}
\end{table}

Tab.~\ref{tab:benchmark_results} summarizes results on the HSI-Benchmark test split under the protocol described in Sec.~\ref{sec:experiments}. 
Despite using only 0.516M trainable parameters (about 2\% of the parameters of the strongest prior backbone), SSFT achieves the best overall performance among all compared methods (84.87). It matches or surpasses substantially larger backbones (e.g., 12M--59M parameters) while improving the overall benchmark score. SSFT attains the best accuracy on Debris (93.33) and the most challenging Fruit domain (61.72), improving upon the previous best Fruit result by 3.44 points (58.28). On HRSS, SSFT remains highly competitive (99.56), close to the best-performing method.

Across a wide range of CNN baselines and recent transformer-based models, SSFT consistently performs strongly across all three domains, indicating robust cross-domain performance on HSI-Benchmark. In particular, the gains on Fruit and Debris, where acquisition conditions differ markedly and class boundaries are defined by subtle spectral cues, suggest that factorized spectral and spatial processing with cross-attention fusion learns representations that remain effective across varied settings. Notably, Fruit remains substantially harder than HRSS and Debris for all methods, highlighting the challenges posed by strong domain shifts and fine-grained spectral differences.


\subsection{Results on SpectralEarth}
\label{subsec:results_spectralearth}

\begin{figure}[t]
   \centering
   \includegraphics[width=.95\columnwidth]{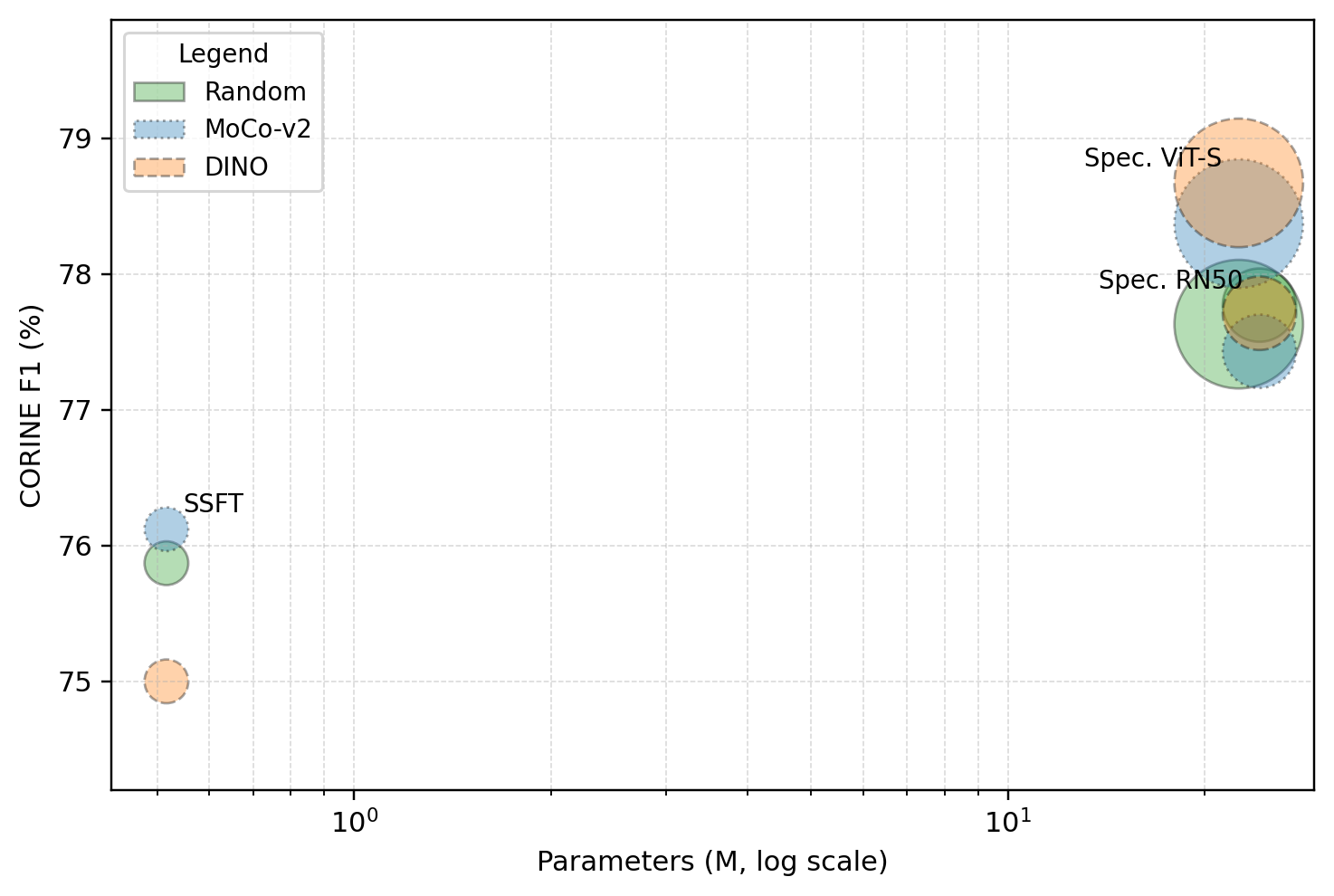}
   \caption{F1-score vs.\@ Parameters and FLOPs on the SpectralEarth~\cite{Braham2025_spectralearth_dataset} CORINE test split. Each point shows the best CORINE F1 achieved for a given weight initialization (Random, MoCo-v2, DINO; best over Frozen vs. Full FT). The x-axis denotes the number of model parameters, while the bubble area is proportional to computational cost (FLOPs).}
   \label{fig:spectralearth_params_flops_f1}
\end{figure}

Tab.~\ref{tab:results_spectral_earth} reports macro F1-score on the SpectralEarth EnMAP-CORINE classification test split following the protocol described in Sec.~\ref{sec:experiments}. Overall, SSFT transfers well to this substantially larger benchmark: fine-tuning (FT) improves performance dramatically compared to linear evaluation across all initializations, and self-supervised initialization provides consistent gains in the frozen setting. In particular, moving from Frozen to Full FT boosts SSFT from 58.89 to 75.87 under Random initialization (+16.98 F1), and MoCo-v2 and DINO further strengthen frozen performance (62.61 and 68.07 F1, respectively), indicating that pretraining yields more transferable features when the backbone is kept fixed.

Compared to significantly larger spectral backbones, SSFT remains competitive under Full FT but does not surpass the strongest large-scale models. In terms of model size, SSFT (0.516M parameters) is substantially more compact than Spec.\ RN50 (24.24M) and Spec.\ ViT-S (22.54M), consistent with Fig.~\ref{fig:spectralearth_params_flops_f1}. Spec.\ ViT-S achieves the best overall results under MoCo-v2 and DINO (78.37 and 78.67 F1), while Spec.\ RN50 is strongest under Random initialization (77.77 F1). The remaining gap suggests that model capacity becomes more important when abundant training data is available, yet SSFT provides a strong trade-off between compactness and performance, particularly when fine-tuned end-to-end. This trade-off is further illustrated by relating the best achieved F1 to model size and computational cost across initialization strategies.

\begin{table}
\caption{Performance on the SpectralEarth \cite{Braham2025_spectralearth_dataset} CORINE classification test split. We compare SSFT against larger spectral backbones under different weight initializations (Random, MoCo-v2, DINO). “Frozen” denotes linear evaluation with a fixed backbone, while “Full FT” fine-tunes all model parameters. Best results for each weight initialization are shown in bold, and second-best results are underlined.}
\label{tab:results_spectral_earth}
\centering
\resizebox{\columnwidth}{!}{%
\begin{tabular}{@{}l l l r@{}}
\toprule
\multirow{2}{*}{Arch.} & \multicolumn{2}{c}{Model Configuration} & \multicolumn{1}{c}{Classif.\ (F1)} \\
\cmidrule(lr){2-3}\cmidrule(l){4-4}
 & Weights & Protocol & \multicolumn{1}{c}{CORINE} \\
\midrule

\multirow{6}{*}{SSFT (ours)} 
& \multirow{2}{*}{Random} 
  & Frozen  & $58.89 \pm 0.51$ \\
& & Full FT & $75.87 \pm 0.41$ \\
\cmidrule(lr){2-4}
& \multirow{2}{*}{MoCo-v2}
  & Frozen  & $62.61 \pm 0.12$ \\
& & Full FT & $76.12 \pm 0.46$ \\
\cmidrule(lr){2-4}
& \multirow{2}{*}{DINO}
  & Frozen  & $68.07 \pm 0.21$ \\
& & Full FT & $75.00 \pm 0.45$ \\
\midrule

\multirow{6}{*}{Spec.\ RN50 \cite{Braham2025_spectralearth_dataset}} 
& \multirow{2}{*}{Random} 
  & Frozen  & $69.92 \pm 0.51$ \\
& & Full FT & \textbf{\boldmath$77.77 \pm 0.38$} \\
\cmidrule(lr){2-4}
& \multirow{2}{*}{MoCo-v2}
  & Frozen  & $74.07 \pm 0.14$ \\
& & Full FT & \underline{$77.43 \pm 0.46$} \\
\cmidrule(lr){2-4}
& \multirow{2}{*}{DINO}
  & Frozen  & $76.61 \pm 0.31$ \\
& & Full FT & \underline{$77.71 \pm 0.55$} \\
\midrule

\multirow{6}{*}{Spec.\ ViT-S \cite{Braham2025_spectralearth_dataset}} 
& \multirow{2}{*}{Random} 
  & Frozen  & $70.59 \pm 0.62$ \\
& & Full FT & \underline{$77.63 \pm 0.37$} \\
\cmidrule(lr){2-4}
& \multirow{2}{*}{MoCo-v2}
  & Frozen  & $73.99 \pm 0.04$ \\
& & Full FT & \textbf{\boldmath$78.37 \pm 0.34$} \\
\cmidrule(lr){2-4}
& \multirow{2}{*}{DINO}
  & Frozen  & $75.11 \pm 0.13$ \\
& & Full FT & \textbf{\boldmath$78.67 \pm 0.34$} \\
\bottomrule
\end{tabular}%
}
\end{table}

\subsection{Ablation Studies}
To further assess our method, we conduct ablations that disable the spectral and spatial streams, evaluate lightweight auxiliary heads, and analyze the impact of common data augmentations on the HSI-Benchmark datasets.

\textbf{Auxiliary Head.} Auxiliary classifier heads provide deep supervision on the branch features prior to fusion. We analyze their effect on the HSI-Benchmark and select the auxiliary loss weight $\lambda_{\mathrm{aux}}$ via a sweep over candidate values on the validation split, using $\lambda_{\mathrm{aux}}=0.05$ in this experiment. Tab.~\ref{tab:aux_head_ablation} shows that auxiliary heads can be beneficial, improving the overall benchmark score by +0.95 (83.92 $\rightarrow$ 84.87) under the same training configuration. However, the effect is domain-dependent: while Debris improves substantially and HRSS increases, performance on Fruit decreases marginally. Overall, these results suggest that deep supervision can strengthen intermediate branch representations, but its impact may vary across domains in a heterogeneous benchmark.

\begin{table}
\caption{Impact of auxiliary classification heads on HSI-Benchmark~\cite{frank2023_hsibenchmark} test accuracy. Best results are shown in bold.}
\label{tab:aux_head_ablation}
\centering
\resizebox{\columnwidth}{!}{%
\begin{tabular}{@{}lcccc@{}}
\toprule
Variant & HRSS & Fruit & Debris & Overall \\
\midrule
SSFT w/o auxiliary heads & 98.48 & \textbf{62.37} & 90.91 & 83.92 \\
SSFT w/ auxiliary heads  & \textbf{99.56} & 61.72 & \textbf{93.33} & \textbf{84.87} \\
\bottomrule
\end{tabular}%
}
\end{table}


\begin{figure*}[t]
   \centering
   \includegraphics[width=0.95\textwidth]{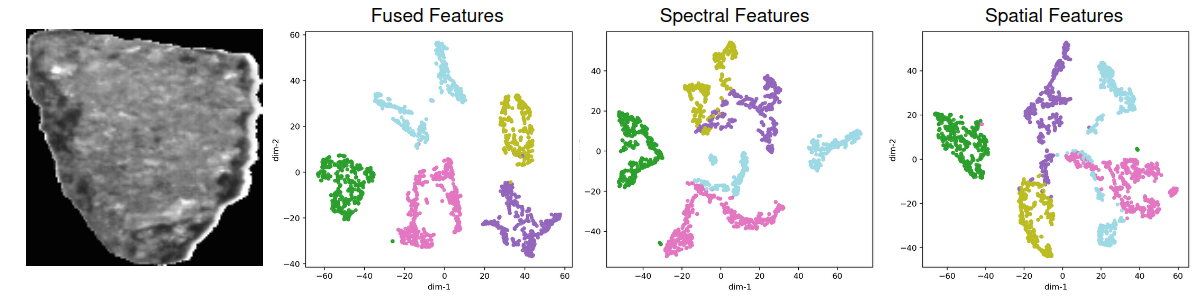}
   \caption{t-SNE visualization of feature representations on the Debris validation split. From left to right, the figure shows a representative input patch from Debris and the 2D t-SNE projections of the fused, spectral, and spatial feature embeddings extracted over the full validation set. The fused representation exhibits a more compact and better separated cluster structure than the individual branch features, suggesting that the two paths encode complementary information.}
   \label{fig:tsne_plot}
\end{figure*}

\textbf{Branch Contributions.} Tab.~\ref{tab:path_ablation} quantifies the contribution of the two branches by disabling one at a time. We set the corresponding feature map to zero before fusion while keeping the rest of the architecture and training setup unchanged, thereby isolating each branch’s contribution at the fusion interface without changing model capacity. 

Disabling the spatial branch ($h^{(p)}\!=\!0$) causes a severe drop in overall accuracy to 38.28, indicating that spatial modeling is essential for robust performance on HSI-Benchmark. Disabling the spectral branch ($h^{(s)}\!=\!0$) is less catastrophic but still reduces the overall score to 70.92, showing that spectral cues provide substantial gains beyond spatial context alone. The largest degradation in this setting occurs on Debris (93.33 $\rightarrow$ 66.46), suggesting that fine-grained spectral information is particularly important for discriminating visually similar materials in that domain. Overall, the full dual-path model consistently outperforms both single-branch variants across all domains, supporting the motivation for factorized spectral-spatial modeling and cross-attention fusion in SSFT; this is also qualitatively reflected in the Debris t-SNE projection in Fig.~\ref{fig:tsne_plot} \cite{maaten2008tsne}, where the fused representation forms a more structured embedding space than the individual branch features.

\begin{table}
\caption{Impact of disabling individual branches on the HSI-Benchmark test split. To disable a branch, we zero out its feature map while keeping the rest of the network unchanged. Best results are shown in bold, and second-best results are underlined.}
\label{tab:path_ablation}
\centering
\resizebox{\columnwidth}{!}{%
\begin{tabular}{@{}lcccc@{}}
\toprule
Variant & HRSS & Fruit & Debris & Overall \\
\midrule
SSFT (full)          & \textbf{99.56} & \textbf{61.72} & \textbf{93.33} & \textbf{84.87} \\
w/o spectral branch  & \underline{90.24} & \underline{56.07} & \underline{66.46} & \underline{70.92} \\
w/o spatial branch   & 21.30 & 43.87 & 49.67 & 38.28 \\
\bottomrule
\end{tabular}%
}
\end{table}


\textbf{Augmentations.} Tab.~\ref{tab:augmentation_ablation} summarizes the effect of common spatial and spectral augmentations on the HSI-Benchmark. Across the tested transformations, training without augmentation yields the best overall performance (84.87), while most augmentations reduce the overall score, with the strongest degradation observed for cropping (82.10 overall). Although some augmentations slightly improve performance on individual domains (\eg, Fruit), these gains do not translate into a higher benchmark score. In particular, several augmentations reduce accuracy on Debris, suggesting that perturbations of spatial structure or spectral signatures can be harmful when classes are visually similar and discrimination relies on fine-grained spectral cues. 

Overall, these results indicate that data augmentation is not a free lunch in diverse hyperspectral settings. Since hyperspectral measurements are constrained by sensor characteristics and the physics of reflectance and illumination, augmentation choices should be validated empirically and, when possible, designed to reflect realistic sources of variation in the target domains. This consideration is particularly important in the small-data regime, where overly aggressive or unrealistic augmentations can easily dominate the limited supervision signal and reduce robustness across settings.

\begin{table}
\caption{Impact of data augmentation on HSI-Benchmark test accuracy. We compare individual spatial and spectral augmentations to training without augmentations. Best results are shown in bold, and second-best results are underlined.}
\label{tab:augmentation_ablation}
\centering
\resizebox{\columnwidth}{!}{%
\begin{tabular}{@{}lcccc@{}}
\toprule
Augmentation & HRSS & Fruit & Debris & Overall \\
\midrule
w/o augmentations      & \textbf{99.56} & 61.72              & \textbf{93.33} & \textbf{84.87} \\
\midrule
flip                   & 98.48          & \underline{62.37}  & 90.91          & 83.92 \\
cut                    & 99.13          & 62.26              & 89.76          & 83.71 \\
rotate                 & 97.99          & 61.27              & \underline{91.33} & 83.53 \\
crop                   & 98.76          & 59.41              & 88.14          & 82.10 \\
\midrule
multiplicative shading & 99.16          & \textbf{62.61}     & 90.44          & \underline{84.07} \\
bandwise gain offset   & \underline{99.28} & 61.69           & 90.35          & 83.77 \\
drop bandblock         & 98.82          & 59.37              & 89.85          & 82.68 \\
wavelength shift       & 99.18          & 61.32              & 89.74          & 83.41 \\
bandwise noise         & 99.21          & 61.95              & 90.97          & 84.05 \\
\bottomrule
\end{tabular}%
}
\end{table}

\section{Conclusion}
\label{sec:conclusion}

We presented SSFT, a lightweight dual-path spectral-spatial fusion transformer for hyperspectral image classification. By factorizing spectral and spatial processing and integrating both streams via cross-attention fusion, SSFT achieves the best overall performance on the multi-domain HSI-Benchmark, spanning three fundamentally different regimes: earth observation (HRSS), close-range fruit condition assessment, and fine-grained material recognition (Debris). These results indicate that SSFT remains effective across benchmark domains despite pronounced differences in sensors, scenes, and acquisition conditions.

Through targeted ablations, we showed that both branches contribute substantially to performance: disabling either the spectral or spatial branch leads to a pronounced drop in accuracy, confirming the benefit of dual-path modeling. The relative impact of each branch varies across domains, indicating that SSFT benefits from complementary spectral and spatial cues under varied acquisition conditions. We further found that auxiliary classification heads can improve overall performance, although their effect is domain-dependent, and that common spatial and spectral data augmentations often reduce robustness in this setting, with training without augmentation performing best.

Beyond HSI-Benchmark, SSFT transfers well to the substantially larger SpectralEarth CORINE benchmark under the official training protocol, providing a strong trade-off between compactness and performance. Overall, our results highlight that robust hyperspectral learning in heterogeneous settings depends not only on architectural choices but also on carefully validated training strategies. Future work may explore domain- and sensor-aware augmentation schemes and more principled objectives that better reflect the physical characteristics of hyperspectral measurements.

 \section*{Acknowledgments}
This work was partially funded by the Carl Zeiss Foundation as part of grant P2024-21-061 under the CZS Transfer program.

{
    \small
    \bibliographystyle{ieeenat_fullname}
    \bibliography{main}
}


\end{document}